\documentclass[a4paper,fleqn]{cas-dc}

\usepackage[numbers]{natbib}
\usepackage{xcolor}
\definecolor{secondblue}{HTML}{3B82C4}
\newcommand{\second}[1]{\textcolor{secondblue}{#1}}
\usepackage{booktabs}
\usepackage{cuted}
\usepackage{caption}
\usepackage{array}
\usepackage{textcomp}
\usepackage{graphicx}

\def\tsc#1{\csdef{#1}{\textsc{\lowercase{#1}}\xspace}}
\tsc{WGM}
\tsc{QE}

\begin{document}
\let\WriteBookmarks\relax
\def\floatpagepagefraction{1}
\def\textpagefraction{.001}

\shorttitle{}    

\shortauthors{Mengmeng Zhang et.al~}  

\title [mode = title]{Multiple Myeloma Lesion Segmentation on Whole-Body Diffusion-Weighted Imaging via Efficient Anatomical Anticipation and Multimodal Confirmation}  



%

\author[1,2]{Mengmeng Zhang}[style=chinese, orcid=0009-0003-8210-8292]
\ead{zhangmengmeng2022@ia.ac.cn}
\credit{Conceptualization of this study, Methodology, Software, Writing – original draft}

\author[3]{Shengqian Huang}[style=chinese, orcid=0000-0003-3257-6082]
\ead{sqhuang18@163.com}
\credit{Data curation, Validation}
\fnmark[1]

\author[3]{Junde Zhou}[orcid=0009-0005-7223-6316]
\ead{zhoujunde@pumch.cn}
\credit{Data curation}

\author[7]{Xiaoping Wu}[orcid=0000-0002-3067-4626]
\ead{xpwu95@163.com}
\credit{Validation}

\author[7,8]{Hao Luo}[orcid=0000-0002-6405-4011]
\ead{michuan.lh@alibaba-inc.com}
\credit{Writing – review, editing}

\author[1,6]{Jing Wang}[orcid=0000-0002-4419-9352]
\ead{wangjing2014@ia.ac.cn}
\credit{Data curation}

\author[1,2]{Yicheng Sun}[orcid=0009-0002-4560-5196]
\ead{sunyicheng2025@ia.ac.cn}
\credit{Data curation}

\author[3,4]{Jiao Li}[orcid=0000-0002-2574-276X]
\ead{ljlyyoung@163.com}
\credit{Data curation}

\author[5]{Haibo Zhang}[orcid=0000-0001-5327-0395]
\ead{zhb\_hello@163.com}
\credit{Data curation}

\author[5]{Sheng Xie}[orcid=0000-0003-3280-9632]
\ead{xs\_mri@126.com}
\credit{Data curation}

\author[7]{Fan Wang}
\ead{fan.w@alibaba-inc.com}
\credit{Resources}

\author[3]{Qin Wang}[orcid=0000-0002-2970-2642]
\ead{wangqin@pumch.cn}
\credit{Resources, Supervision, Funding acquisition}
\cormark[1]

\author[3]{Huadan Xue}[orcid=0000-0002-4278-2165]
\ead{xuehd@pumch.cn}
\credit{Resources, Supervision, Funding acquisition}
\cormark[1]

\author[1,2]{Yisheng Lv}[orcid=0000-0002-0508-1298]
\ead{yisheng.lv@ia.ac.cn}
\credit{Resources, Supervision}
\cormark[1]

\author[1]{Fei-yue Wang}
\ead{feiyue.wang@ia.ac.cn}
\credit{ Supervision}
\cormark[1]

\affiliation[1]{organization={the State Key Laboratory of Multimodal Artificial Intelligence Systems, Institute of Automation, Chinese Academy of Sciences},
            city={Beijing},
            postcode={100190}, 
            state={Beijing},
            country={China}}

\affiliation[2]{organization={the School of Artificial Intelligence, University of Chinese Academy of Sciences},
            city={Beijing},
            postcode={101408}, 
            state={Beijing},
            country={China}}

\affiliation[3]{organization={the Department of Radiology, Peking Union Medical College Hospital, Chinese Academy of Medical Sciences \& Peking Union Medical College},
            city={Beijing},
            postcode={100730}, 
            state={Beijing},
            country={China}}

\affiliation[4]{organization={the Department of Medical Imaging, Peking University Shenzhen Hospital},
            city={Shenzhen},
            postcode={518036}, 
            state={Guangdong},
            country={China}}

\affiliation[5]{organization={the Department of Radiology, China-Japan Friendship Hospital},
            city={Beijing},
            postcode={100029}, 
            state={Beijing},
            country={China}}

\affiliation[6]{organization={the Faculty of Innovation Engineering, Macau University of Science and Technology},
            city={Macao},
            postcode={999078}, 
            state={Macao},
            country={China}}

\affiliation[7]{organization={the DAMO Academy, Alibaba Group},
            city={Hangzhou},
            state={Zhejiang},
            country={China}}

\affiliation[8]{orgnization={the Hupan Lab, Zhejiang Province},
                city={Hangzhou},
                state={Zhejiang},
                country={China}}

\cortext[1]{Corresponding author}

\fntext[1]{This author contributed equally to this work as M.Zhang}


\begin{abstract}
Whole-body diffusion-weighted imaging (WB-DWI) is widely used for multiple myeloma (MM) assessment, yet automated lesion segmentation remains challenging due to limited anatomical delineation and the low specificity of marrow hyperintensity. Existing studies have introduced bone region-of-interest (ROI) information and apparent diffusion coefficient (ADC) maps to mitigate these ambiguities, but practical limitations remain. Bone ROI construction often relies on costly manual annotation, image registration, or dedicated bone models, while ADC is usually incorporated only through simple channel fusion, limiting its ability to provide complementary structural and lesion-discriminative cues.
To address these limitations, we propose a two-stage framework for MM lesion segmentation on WB-DWI. In the first stage, we train a bone ROI generation model from ADC images without dedicated bone labels, providing an efficient and practical anatomical prior for lesion analysis. In the second stage, we propose Anatomy-guided Multimodal U-Net (AMU-Net), which leverages ADC in a manner consistent with clinical lesion assessment rather than treating it as a generic auxiliary modality.
Extensive experiments demonstrate the effectiveness and practicality of the proposed method. It achieves the best overall performance among the evaluated methods, with a mean Dice score of 76.2\%.
\end{abstract}





\maketitle

\section{Introduction}

Whole-body diffusion-weighted imaging (WB-DWI) is widely used for diagnosis and response assessment in multiple myeloma (MM) because it provides whole-body coverage without ionizing radiation or contrast administration~\cite{summers2021whole,messiou2015whole}. Automated lesion segmentation on WB-DWI would be clinically valuable, but remains challenging. Although WB-DWI highlights diffusion-restricted lesions, it provides limited anatomical delineation of the skeleton, making it difficult to distinguish marrow lesions from adjacent soft tissue or other extraskeletal structures. In addition, marrow hyperintensity is not specific to MM and may also arise from benign marrow hyperplasia, inflammation, edema, and other nonmalignant processes.

In clinical reading, suspicious hyperintensity on WB-DWI is therefore not interpreted in isolation. Radiologists rely on both anatomical context and ADC maps derived from the same diffusion acquisition. Anatomical knowledge helps determine whether suspicious signal is plausibly located within marrow-containing bone, while ADC provides complementary tissue characterization for lesion assessment~\cite{paternain2020utility,torkian2023diffusion,baliyan2016diffusion}. Motivated by this reading strategy, previous studies have introduced bone ROI information and ADC into MM lesion segmentation pipelines~\cite{bauer2025advanced,giles2015assessing}. However, important limitations remain. Bone ROI construction often relies on manual annotation, atlas-based registration, or dedicated bone models~\cite{wennmann2023deep,candito2025weakly,candito2024deep}, all of which increase cost and complexity. Meanwhile, ADC is typically incorporated only through simple input-level concatenation, without explicitly modeling its distinct roles in structural supplementation and lesion confirmation.

To address these limitations, we propose a two-stage framework for MM lesion segmentation on WB-DWI. In the first stage, we train a bone ROI generation model from inherently co-registered ADC images without dedicated bone labels, producing an anatomical prior for downstream lesion segmentation. In the second stage, we propose Anatomy-guided Multimodal U-Net (AMU-Net), which leverages ADC in a manner consistent with clinical lesion assessment. Specifically, ADC is used to supplement anatomical structure at the encoder and to support lesion confirmation at the decoder under explicit ROI guidance.  Our main contributions can be summarized as follows:
\begin{itemize}
    \item We develop a bone ROI generation strategy that learns skeletal priors from ADC images without dedicated bone labels, avoiding additional bone annotation and registration.
    \item We propose Anatomy-guided Multimodal U-Net (AMU-Net) for MM lesion segmentation on WB-DWI, which uses ADC for both anatomy-aware structural supplementation and lesion confirmation under ROI guidance.
    \item We validate the proposed design through systematic experiments and ablation studies, showing that annotation-efficient ROI generation and clinically motivated ADC utilization lead to improved MM lesion segmentation performance. 
\end{itemize}
    
To facilitate community use, we will publicly release the code repository and trained model weights upon acceptance at \url{https://github.com/M-MZhang/MM_segmentation.git}.

\begin{figure}
    \centering
    \includegraphics[width=\columnwidth]{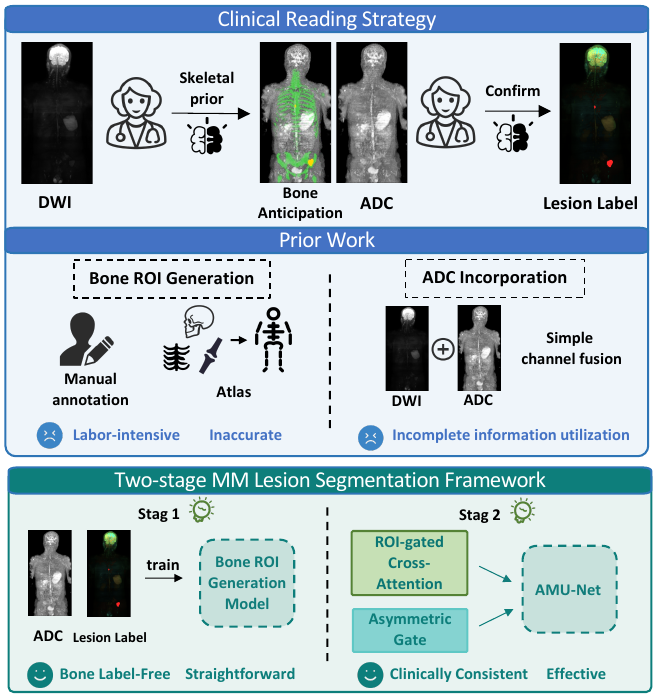}
    \caption{\textbf{Motivation of the proposed framework.} In clinical reading, radiologists first anticipate anatomically plausible marrow regions as skeletal priors and then jointly interpret WB-DWI and ADC to confirm MM lesions. Prior work addresses these cues only partially, relying on manual or atlas-based bone ROI generation and simple ADC channel fusion. Inspired by this strategy, our method adopts a two-stage design: Stage 1 learns a bone ROI generation model from ADC without dedicated bone labels, and Stage 2 uses AMU-Net for anatomy-guided multimodal lesion segmentation.}
    \label{fig:motivation}
\end{figure}

\section{Related work}



\begin{figure*}[t]
    \centering
    \includegraphics[width=\textwidth]{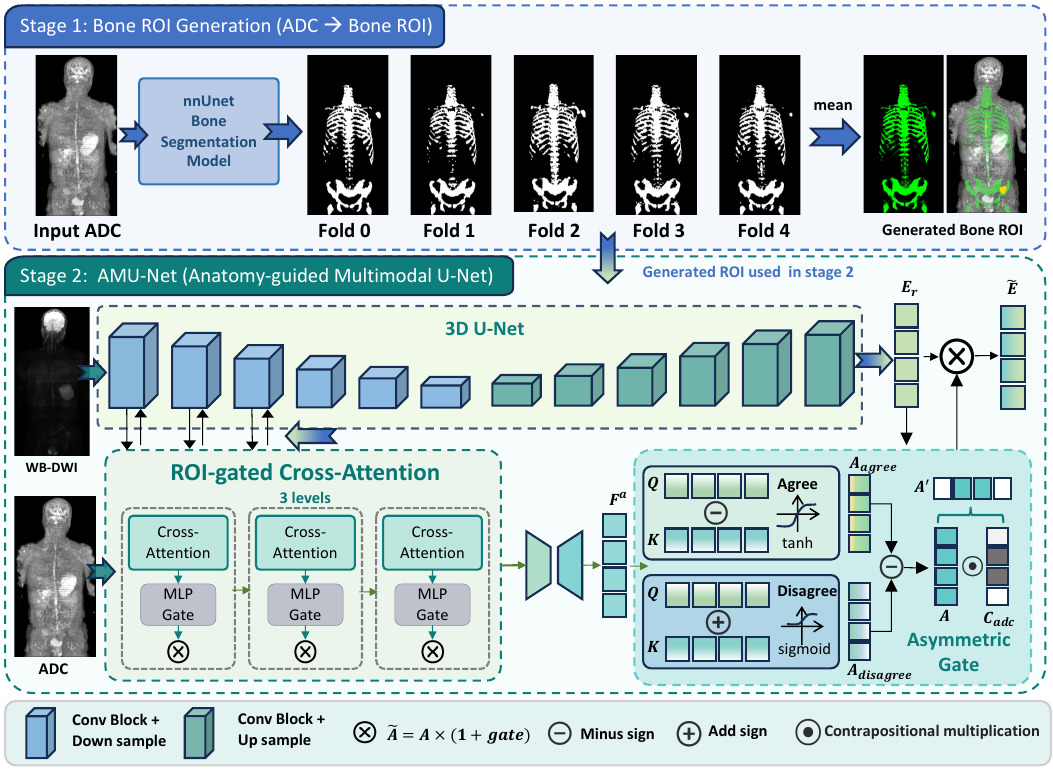}
    \caption{\textbf{Overview of the proposed two-stage framework for MM lesion segmentation on WB-DWI.} Stage I generates bone ROI maps from ADC images using a 3D nnUNet ensemble. Stage II uses AMU-Net to perform ROI-guided lesion segmentation from WB-DWI with ADC assistance, via encoder-level ROI-gated cross-attention and a decoder-level asymmetric gate.}
    \label{fig:model-structure}
    \vspace{-15pt}
\end{figure*}

\subsection{Bone ROI Acquisition and Anatomy-Aware Segmentation}

Previous studies have recognized that bone or bone marrow region-of-interest (ROI) information can serve as an effective anatomical prior for multiple myeloma (MM) lesion segmentation, as it helps suppress false positives and constrains predictions to plausible marrow-bearing regions~\cite{wuts2026clinically}. This has motivated a line of work on automatic bone or bone marrow ROI acquisition~\cite{michoux2021repeatability,tiribilli2024single, nguyen2016towards, varghese2018transforming,candito2025weakly}. However, most such methods have been developed on modalities with clearer structural depiction, such as CT or anatomical MRI, rather than on WB-DWI, which is more commonly used for lesion assessment in MM~\cite{koutoulakis2025label,bauer2025advanced}. On WB-DWI, the weak delineation of skeletal and marrow boundaries makes both manual annotation and direct structure learning substantially more challenging. Consequently, methods developed on other modalities are not readily applicable to WB-DWI-based lesion segmentation. The relatively few attempts to obtain skeletal ROIs directly on WB-DWI have mainly relied on atlas- or registration-based localization~\cite{candito2025weakly}, which is often time-consuming and sensitive to variations in patient positioning, field-of-view, and image appearance, thereby limiting its robustness in practice.

\subsection{ADC-Guided Multimodal Fusion for MM Lesion Segmentation}


Apparent diffusion coefficient (ADC) maps are a natural auxiliary input for multiple myeloma (MM) lesion segmentation because they are derived from the same WB-DWI acquisition and provide quantitative diffusion-related information~\cite{horger2011whole,bonaffini2015apparent,wu2018discriminating}. Since ADC maps are inherently aligned with WB-DWI and reflect tissue cellularity, they can provide complementary evidence for distinguishing true pathological diffusion restriction from normal marrow heterogeneity. Existing MM methods that use ADC typically adopt simple input-level fusion by concatenating ADC with WB-DWI as an additional channel~\cite{almeida2020quantification}. While effective as a straightforward baseline, this strategy treats ADC as a generic secondary modality and does not explicitly capture its task-specific role in MM, where it may assist both lesion confirmation and structural delineation within the marrow. More advanced multimodal fusion methods, such as feature-wise modulation, gated fusion, and attention-based interaction, have been explored in broader medical image segmentation settings~\cite{chen2024modality,mo2021mutual,zhuang20243d,zhuang20223d}, but they are mainly evaluated on modality pairs such as multiparametric MRI or CT/PET. Their suitability for MM lesion segmentation on WB-DWI therefore remains unclear.

\section{Method}

\subsection{Overview}

The method is organized as a two-stage framework for multiple myeloma (MM) lesion segmentation on whole-body diffusion-weighted imaging (WB-DWI). Bone ROI maps are first generated from ADC images, after which lesion segmentation is performed by ADC-guided Multimodal U-Net (AMU-Net). The overall architecture comprises three key components: (i) annotation-efficient bone ROI generation, (ii) encoder-level ROI-gated ADC--DWI interaction for structural supplementation, and (iii) decoder-level asymmetric gate for lesion confirmation, as shown in Fig.~\ref{fig:model-structure}.

\subsection{Stage I: Annotation-Efficient Bone ROI Generation}

To achieve annotation-efficient bone ROI generation, we exploit a subset of diffuse MM cases in which lesion involvement spans most of the marrow cavity. In these cases, the available lesion annotations can serve as surrogate supervision for skeletal marrow extent, thereby avoiding additional manual bone annotation. Combined with the clearer structural delineation provided by ADC compared with WB-DWI, this enables learning of a task-specific ROI generator for downstream lesion segmentation.

Specifically, let \(A \in \mathbb{R}^{H \times W \times D}\) denote the ADC volume, and let \(R \in [0,1]^{H \times W \times D}\) denote the predicted bone ROI probability map. We train a 3D nnUNet ensemble on ADC using lesion labels from diffuse MM cases as pseudo-marrow supervision. The goal of this model is not precise anatomical bone segmentation, but rather generation of a soft ROI prior with sufficient skeletal coverage for subsequent lesion segmentation. Each ensemble member \(m=1,\dots,M\) predicts a probability map \(R^{(m)}\), and the final ROI is obtained by voxel-wise mean aggregation:
\begin{equation}
R(\mathbf{v}) = \frac{1}{M} \sum_{m=1}^{M} R^{(m)}(\mathbf{v}),
\end{equation}
where \(\mathbf{v}\) indexes voxel location. This aggregation reduces prediction variance across ensemble members and produces a more stable soft anatomical prior for Stage II.

Because ADC is acquired in the same diffusion sequence as DWI, the ROI map is inherently aligned with the lesion segmentation input and can be used directly without additional registration.

\subsection{Stage II: AMU-Net}
Stage II performs lesion segmentation on WB-DWI under explicit bone ROI guidance from Stage I. AMU-Net, is built on a 3D U-Net backbone with DWI as the primary input for lesion representation and ADC as an auxiliary modality. Guided by the clinical practice of jointly interpreting DWI and ADC, AMU-Net incorporates ADC information in two complementary ways. At the encoder level, ROI-gated cross-attention is used to inject structural cues from ADC into DWI features, compensating for the limited anatomical delineation of WB-DWI. At the decoder level, a decoder-level asymmetric gate modulates the final lesion embedding according to DWI--ADC concordance for lesion confirmation. Together, these designs enable ROI- and ADC-aware lesion segmentation while preserving the lesion-sensitive representation learned from DWI.

\subsubsection{Encoder-level ROI-gated cross-attention}

WB-DWI enhances lesion conspicuity by suppressing background signal, but this also weakens anatomical depiction of the skeleton. To supplement the missing structural information, we introduce ROI-gated ADC--DWI cross-attention at multiple encoder levels.

Let \(F_l^d\) and \(F_l^a\) denote the DWI and ADC feature maps at encoder level \(l\), respectively, and let \(R_l\) denote the Stage-I ROI map resized to the same spatial resolution. For each selected level, the two feature maps are concatenated and passed through a lightweight gating function:
\begin{equation}
G_l = \sigma \left( \phi_l([F_l^d, F_l^a]) \right) \odot R_l
\end{equation}
where \([\,\cdot,\cdot\,]\) denotes channel-wise concatenation, \(\phi_l(\cdot)\) is a two-layer \(1\times1\times1\) convolutional projection with ReLU activation, \(\sigma(\cdot)\) denotes the sigmoid function, and \(\odot\) denotes element-wise multiplication.

The DWI feature is then modulated as
\begin{equation}
\tilde{F}_l^d = F_l^d \odot (1 + G_l)
\end{equation}

This design allows ADC to supplement DWI features with clearer structural cues while preserving DWI as the primary lesion-sensitive representation. The ROI gate further constrains cross-modal interaction to likely skeletal regions, reducing interference from extra-skeletal tissue. We apply this interaction at the first three encoder levels, as shallow encoder features preserve richer spatial and anatomical detail, making them more suitable for ADC-based structural supplementation than deeper semantic features.

\begin{figure*}
    \centering
    \includegraphics[width=\textwidth]{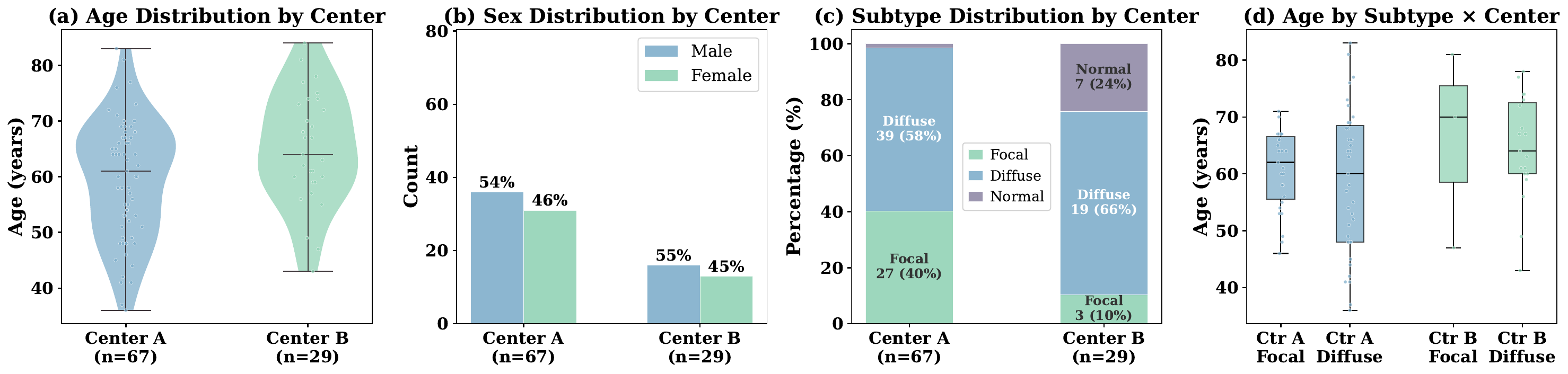}
    \caption{\textbf{Demographic and clinical characteristics of the internal and external cohorts:} (a) age distribution by center, (b) sex distribution by center, (c) subtype distribution by center, and (d) age distribution stratified by subtype and center. Blue and green denote the internal and external cohorts, respectively. Violin plots show density distributions with embedded boxplots.}
    \label{fig:demographics}
\end{figure*}

\begin{figure*}
    \centering
    \includegraphics[width=\textwidth]{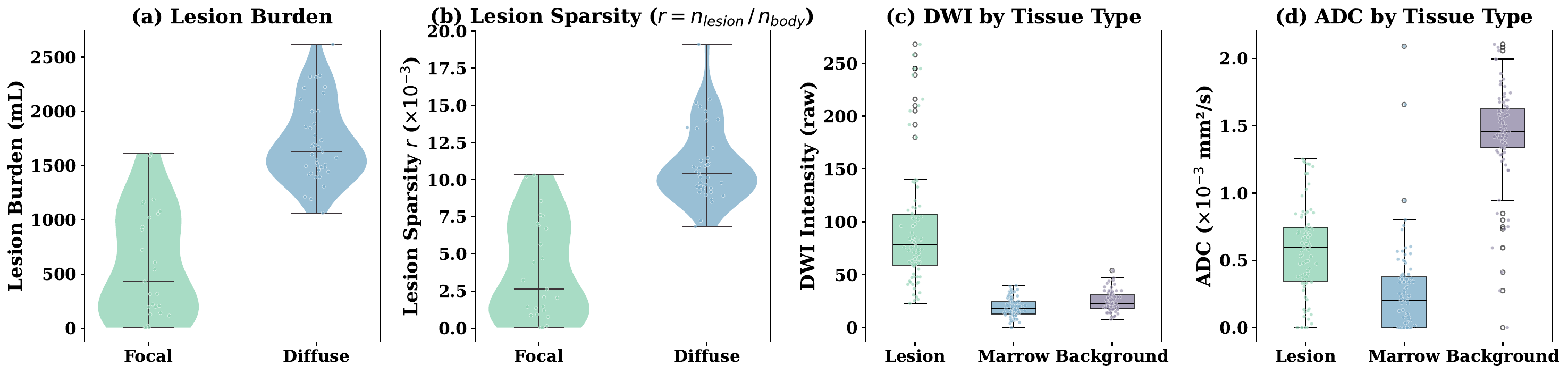}
   \caption{\textbf{Task-related data characteristics of the internal cohort:} (a) lesion burden, (b) lesion sparsity, (c) DWI signal distributions by tissue type, and (d) ADC distributions by tissue type. Green and blue denote focal and diffuse cases in (a) and (b), while colors in (c) and (d) indicate lesion, marrow, and background tissues. Violin plots show density distributions with embedded boxplots, and dots represent individual observations.}
    \label{fig:data_characteristics}
    \vspace{-12pt}
\end{figure*}

\subsubsection{Decoder-level asymmetric gate}

Within the marrow cavity, suspicious DWI hyperintensity is not specific to MM lesions and is often interpreted together with ADC in clinical reading. To further improve lesion confirmation, we introduce a decoder-level asymmetric gate that modulates the final DWI embedding according to DWI--ADC concordance and discrepancy.

Let \(E \in \mathbb{R}^{C \times H \times W \times D}\) denote the final full-resolution DWI decoder embedding, and let \(F^a \in \mathbb{R}^{C_a \times H \times W \times D}\) denote the corresponding full-resolution ADC feature map. We first inject the explicit skeletal prior by ROI-based amplification:
\begin{equation}
E_r = E \odot (1 + R)
\end{equation}
where the ROI map \(R\) is broadcast along the channel dimension.

Next, point-wise linear projections map the DWI and ADC features into a shared latent space:
\begin{equation}
Q = W_q(E_r) \qquad K = W_k(F^a)
\end{equation}
where \(W_q\) and \(W_k\) are learnable \(1\times1\times1\) convolutions.

The agree branch captures joint DWI--ADC evidence:
\begin{equation}
A_{\mathrm{agree}} = \tanh \left( \psi_a(Q + K) \right)
\end{equation}
where \(\psi_a(\cdot)\) is a \(1\times1\times1\) convolution producing a single-channel response map. The \(\tanh\) activation allows this branch to either enhance or suppress the embedding depending on the local joint DWI--ADC pattern.

The disagree branch captures discrepant evidence between DWI and ADC:
\begin{equation}
A_{\mathrm{disagree}} = \sigma \left( \psi_d(Q - K) \right)
\end{equation}
where \(\psi_d(\cdot)\) is another \(1\times1\times1\) convolution and \(\sigma(\cdot)\in[0,1]\). This branch acts as a one-sided suppressive term for potentially misleading cross-modality discrepancy.

The combined attention map is defined as
\begin{equation}
A = A_{\mathrm{agree}} - A_{\mathrm{disagree}}
\end{equation}

To reduce the influence of unreliable ADC responses, we further introduce a simple ADC confidence term \(C_{\mathrm{adc}}\). Although ADC values are expected to be non-negative by definition, voxel-wise ADC measurements may still be affected by the long WB-DWI acquisition process and related noise, distortion, and motion artifacts. We therefore define \(C_{\mathrm{adc}}\) based on whether the local ADC value is positive:
\begin{equation}
C_{\mathrm{adc}}(x) =
\begin{cases}
1, & \mathrm{ADC}(x) > 0 \\
0, & \text{otherwise}
\end{cases}
\end{equation}
and use it to restrict the attention modulation to locations with valid ADC support:
\begin{equation}
A' = A \odot C_{\mathrm{adc}}
\end{equation}

The refined decoder embedding is then computed as
\begin{equation}
\tilde{E} = E_r + A' \odot E_r
\end{equation}

This asymmetric design distinguishes concordance from discrepancy in DWI--ADC interaction, while the confidence term limits ADC-guided modulation to locations with valid ADC support. The refined embedding \(\tilde{E}\) is finally fed to the classification head to produce lesion logits.

\subsection{Training Objectives}

The two stages are trained separately. Stage I follows the standard nnUNet training setup for ROI prediction. After training, the Stage-I model is fixed and used to generate ROI probability maps for Stage II.

For lesion segmentation, let \(Y\) and \(Y^\ast\) denote the predicted and reference lesion masks, respectively. Stage II is optimized using a combination of Dice loss and binary cross-entropy (BCE) loss:
\begin{equation}
\mathcal{L}_{\mathrm{seg}} = \lambda_1 \mathcal{L}_{\mathrm{Dice}}(Y, Y^\ast) + \lambda_2 \mathcal{L}_{\mathrm{BCE}}(Y, Y^\ast)
\end{equation}
where \(\lambda_1\) and \(\lambda_2\) are balancing weights. Unless otherwise specified, both are set to 1.
\begin{table*}
\centering
\caption{Internal comparison with baseline models under the same 5-fold evaluation protocol. Best results are shown in \textbf{bold}, and second-best results are shown in \second{blue}.}
\label{tab:main_results}
\setlength{\tabcolsep}{6pt}
\renewcommand{\arraystretch}{1.08}
\resizebox{\textwidth}{!}{
\begin{tabular}{@{}l c c c c c c c@{}}
\toprule
Model & \multicolumn{1}{c}{Type} & Params & Dice (\%) \second{$\uparrow$} & IoU (\%) \second{$\uparrow$} & HD95 \second{$\downarrow$} & Precision (\%) \second{$\uparrow$} & Recall (\%) \second{$\uparrow$}\\
\midrule

nnU-Net\cite{isensee2021nnu} & \multirow{6}{*}{\parbox[c]{1.5cm}{\centering\scriptsize 3D CNN}} 
& $\sim$30M  & \second{$73.13 \pm 1.5 $} & \second{$63.24 \pm 1.3$} & \second{$11.8 \pm 1.5$} & $\mathbf{80.0}$ & 76.3 \\

VISTA3D\cite{he2025vista3d} & & $\sim$30M  & $72.40 \pm 0.4$ & $61.43 \pm 0.7$ & $12.6 \pm 0.5$ & 73.0 & $\mathbf{82.3}$ \\

SegResNet\cite{myronenko20183d} & & $\sim$19M & $67.04 \pm 0.7$ & $56.00 \pm 0.7$ & $16.3 \pm 0.8$ & 67.3 & 76.7 \\

U-Net\cite{ronneberger2015u} & & $\sim$5M & $64.92 \pm 1.4$ & $52.93 \pm 1.5$ & $16.5 \pm 0.5$ & 68.3 & 73.5 \\

DynUNet\cite{cardoso2022monai} & & $\sim$5M & $63.81 \pm 1.7$ & $51.93 \pm 1.9$ & $16.7 \pm 1.3$ 
& 65.8 & 74.0 \\

AttentionUNet\cite{oktay2018attention} & & $\sim$5M & $63.29 \pm 0.5$ & $51.45 \pm 0.8$ & $16.2 \pm 0.7$ & 65.5 & 76.1 \\
\midrule

SwinUNETR\cite{hatamizadeh2021swin} & \multirow{2}{*}{\parbox[c]{1.5cm}{\centering\scriptsize 3D\\Transformer}} & $\sim$62M  & $63.72 \pm 0.4$ & $52.11 \pm 0.6$ & $19.4 \pm 0.5$ & 64.6 & 74.8 \\

UNETR\cite{hatamizadeh2022unetr} & & $\sim$93M  & $57.32 \pm 1.1$ & $45.09 \pm 1.1$ & $24.3 \pm 1.0$ 
& 53.2 & 78.1 \\
\midrule

MA-Net\cite{fan2020ma} & \multirow{4}{*}{\parbox[c]{1.5cm}{\centering\scriptsize 2D CNN}} & $\sim$28M  & $72.75 \pm 1.4$ & $61.96 \pm 1.6$ & $11.9 \pm 1.4$ & 71.4 & \second{$82.1$} \\

UNeXt\cite{valanarasu2022unext} & & $\sim$1.5M & $71.67 \pm 0.2$ & $60.12 \pm 0.3$ & $12.0 \pm 0.7$ & 70.6 & 81.0 \\

MultiResUNet\cite{ibtehaz2020multiresunet} & & $\sim$7M   & $68.32 \pm 1.5$ & $56.02 \pm 1.9$ & $13.1 \pm 1.3$ & 65.5 & 79.0 \\

MMNet\cite{zhao2024multiple} & & $\sim$1.5M & $68.46 \pm 3.3$  & $56.70 \pm 4.3$  & $15.2 \pm 3.0$ & 68.1 & 77.9 \\
\midrule

\textbf{AMU-Net(Ours)} & -- & $\sim$31M  & $\mathbf{76.24 \pm 0.5}$ & $\mathbf{65.93 \pm 1.0}$ & $\mathbf{10.1 \pm 0.7}$ & \second{$78.3$} & \second{$82.1$} \\
\bottomrule
\end{tabular}
}
\end{table*}

\begin{table*}
\centering
\caption{Zero-shot external validation on an independent cohort. Models trained using the same 5-fold protocol were directly tested on the external dataset without any fine-tuning or adaptation. Best results are shown in \textbf{bold}, and second-best results are shown in \second{blue}.}
\label{tab:external_results}
\setlength{\tabcolsep}{6pt}
\renewcommand{\arraystretch}{1.08}
\resizebox{\textwidth}{!}{
\begin{tabular}{@{}l c c c c c c c@{}}
\toprule
Model & \multicolumn{1}{c}{Type} & Params & Dice (\%) \second{$\uparrow$} & IoU (\%) \second{$\uparrow$} & HD95 \second{$\downarrow$} & Precision (\%) \second{$\uparrow$} & Recall (\%) \second{$\uparrow$} \\
\midrule

nnU-Net\cite{isensee2021nnu}
& \multirow{6}{*}{\parbox[c]{1.5cm}{\centering\scriptsize 3D CNN}}
& $\sim$30M
& $28.70 \pm 1.3$
& $18.59 \pm 1.1$
& 70.29 $\pm$ 0.82
& 32.31
& 35.16 \\

VISTA3D\cite{he2025vista3d}
&
& $\sim$30M
& $30.83 \pm 1.1$
& $21.26 \pm 0.9$
& 61.49 $\pm$ 0.79
& 26.44
& 48.78 \\

SegResNet\cite{myronenko20183d}
&
& $\sim$19M
& \second{$40.65 \pm 0.9$}
& \second{$29.96 \pm 0.8$}
& \second{53.41 $\pm$ 5.02}
& \second{44.79}
& 44.93 \\

U-Net\cite{ronneberger2015u}
&
& $\sim$5M
& $39.64 \pm 2.7$
& $28.97 \pm 2.5$
& 63.98 $\pm$ 3.02
& 36.99
& 53.28 \\

DynUNet\cite{cardoso2022monai}
&
& $\sim$5M
& $39.42 \pm 2.4$
& $28.55 \pm 2.2$
& 62.20 $\pm$ 4.46
& 35.43
& 56.46 \\

AttentionUNet\cite{oktay2018attention}
&
& $\sim$5M
& $35.99 \pm 1.4$
& $25.58 \pm 1.2$
& 69.05 $\pm$ 6.72
& 32.46
& $\mathbf{57.80}$ \\
\midrule

SwinUNETR\cite{hatamizadeh2021swin}
& \multirow{2}{*}{\parbox[c]{1.5cm}{\centering\scriptsize 3D\\Transformer}}
& $\sim$62M
& $35.43 \pm 1.8$
& $24.99 \pm 1.7$
& 66.33 $\pm$ 2.87
& 32.26
& 50.30 \\

UNETR\cite{hatamizadeh2022unetr}
&
& $\sim$93M
& $30.77 \pm 0.5$
& $20.74 \pm 0.4$
& 68.39 $\pm$ 4.51
& 25.06
& 54.22 \\
\midrule

MA-Net\cite{fan2020ma}
& \multirow{4}{*}{\parbox[c]{1.5cm}{\centering\scriptsize 2D CNN}}
& $\sim$28M
& $37.26 \pm 1.7$
& $27.62 \pm 1.5$
& 50.47 $\pm$ 5.10
& 39.79
& 41.26 \\

UNeXt\cite{valanarasu2022unext}
&
& $\sim$1.5M
& $38.11 \pm 2.5$
& $28.00 \pm 2.1$
& 56.03 $\pm$ 1.48
& 35.88
& 51.10 \\

MultiResUNet\cite{ibtehaz2020multiresunet}
&
& $\sim$7M
& $37.86 \pm 2.6$
& $27.28 \pm 2.1$
& \textbf{46.34 $\pm$ 7.34}
& $44.69$
& 37.77 \\

MMNet\cite{zhao2024multiple}
&
& $\sim$1.5M
& $30.66 \pm 2.9$
& $22.11 \pm 2.4$
& 57.56 $\pm$ 1.24
& 28.48
& \textbf{49.05} \\
\midrule

\textbf{AMU-Net(Ours)}
& --
& $\sim$31M
& \textbf{43.43 $\pm$ 0.9}
& \textbf{33.04 $\pm$ 0.9}
& 59.93 $\pm$ 3.31
& \textbf{48.00}
& \second{47.43} \\
\bottomrule
\end{tabular}
}
\vspace{-10pt}
\end{table*}

\section{Experiments}
\subsection{Dataset and Data Analysis}

\subsubsection{Cohort Description}

This study used two whole-body MRI cohorts collected from different institutions. The internal cohort, acquired at \textbf{Peking Union Medical College Hospital}, comprised 67 cases. Among them, 47 cases were used for model development and were further evaluated under a 5-fold cross-validation protocol, while the remaining 20 cases were reserved as an independent internal validation set. The external cohort, acquired at \textbf{China-Japan Friendship Hospital}, comprised 29 cases and was reserved for independent external validation. The distributions of MM subtypes, age, and sex in the two cohorts are summarized in Fig.~\ref{fig:demographics}.

\subsubsection{Lesion Distribution Characteristics}

To characterize the complexity of whole-body MM lesion segmentation, we quantified lesion burden and lesion sparsity on the internal cohort, as shown in Fig.~\ref{fig:data_characteristics}(a)--(b). Lesion burden was defined as the total lesion volume per patient, measured in milliliters from manual voxel-wise annotations. Lesion sparsity was defined as the ratio of lesion-positive voxels to body voxels, i.e., \(r = n_{\text{lesion}} / n_{\text{body}}\), reflecting how sparsely lesions are distributed within the whole-body field of view. Body voxels were obtained from the foreground body region after preprocessing, and lesion voxels were counted from the corresponding expert annotations. These statistics show that diffuse cases generally exhibited substantially larger lesion burden and higher lesion occupancy than focal cases. At the same time, the lesion-positive ratio remained extremely low overall, indicating severe class imbalance and sparse target distribution.

\begin{figure}
    \centering
    \includegraphics[width=1\linewidth]{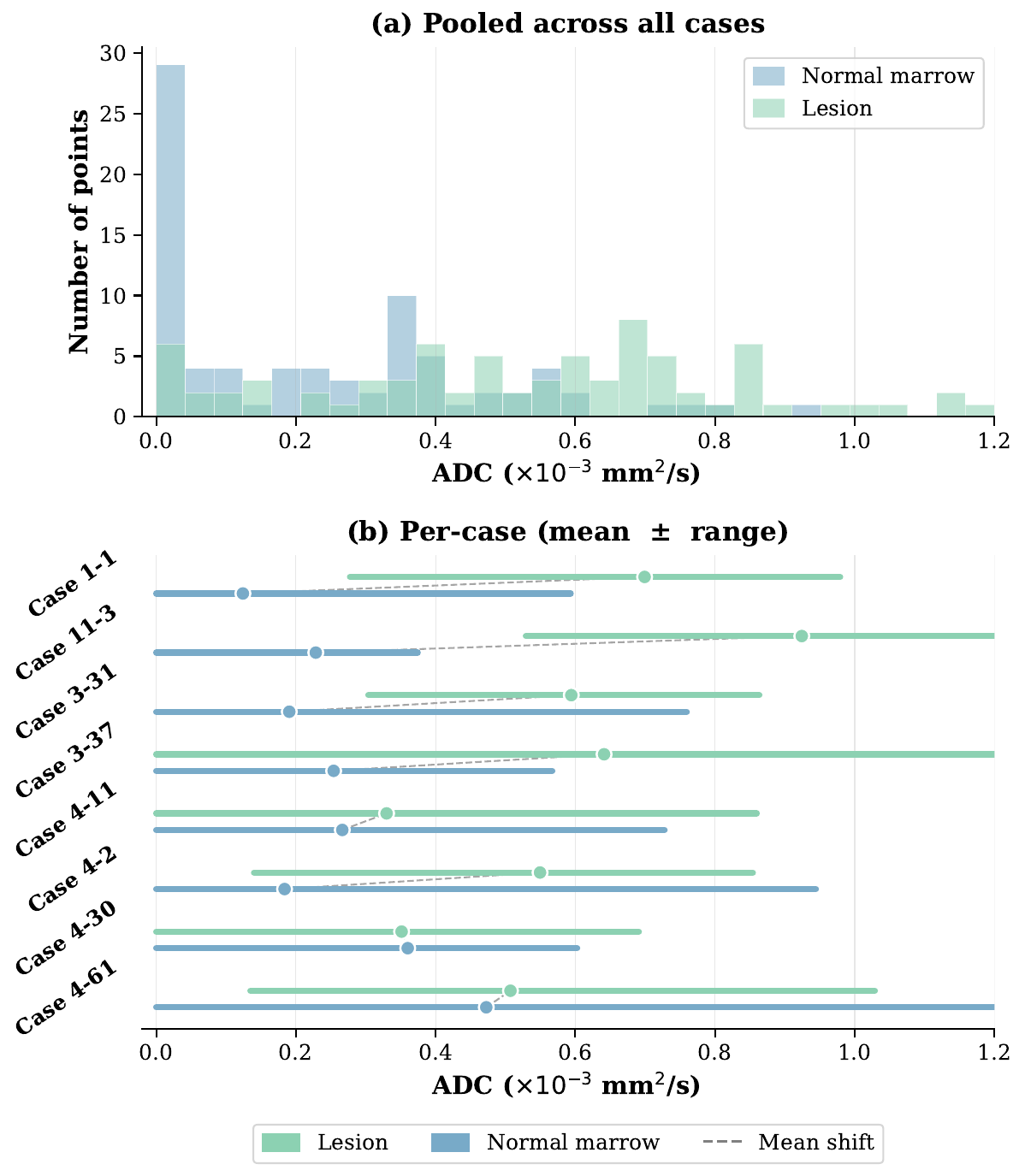}
    \caption{\textbf{ADC distributions of lesion and normal marrow from pooled and per-case perspectives.} (a) Pooled across all cases, lesion and normal marrow ADC values substantially overlap, suggesting that absolute ADC values alone are insufficient for disease assessment. (b) Per-case mean and range show that lesion-related abnormalities are more meaningfully assessed relative to the patient's own marrow background. This supports the role of bone ROI guidance in constraining DWI--ADC interpretation to the skeletal marrow space.}
    \label{fig:adc_two_views}
    \vspace{-10pt}
\end{figure}

\subsubsection{DWI and ADC Signal Characteristics}

We further analyzed modality-specific tissue signals using 247 radiologist-annotated points from 8 selected cases, covering lesion, marrow, and background regions (Fig.~\ref{fig:data_characteristics}(c)--(d)). For each annotated point, the corresponding DWI and ADC values were sampled from the original images to examine tissue-level signal distributions. The resulting plots show distinct yet partially overlapping signal patterns across tissue types in both modalities, indicating that DWI and ADC provide complementary but individually insufficient information for lesion identification. Overall, these analyses highlight several key challenges of MM lesion segmentation on WB-DWI, including large inter-patient variation in lesion extent, extremely sparse lesion occupancy, and heterogeneous multimodal signal characteristics.

\subsection{Experimental Setup}
To evaluate the effectiveness and generalizability of the proposed method, all primary experiments were performed using a patient-level 5-fold cross-validation strategy. Unlike sample-level splitting, patient-level partitioning ensures that all images or records associated with the same patient are assigned to the same fold, thereby preventing information leakage between the training and test sets. The dataset was randomly divided into five mutually exclusive folds at the patient level. For each iteration, four folds were used for training and internal validation, and the remaining fold was used for testing. This process was repeated five times, with each fold used once as the test set.

All compared methods were evaluated under the same cross-validation protocol and data splits to ensure a fair comparison. The final quantitative results were reported as the mean and standard deviation across the five folds, providing an estimate of both the overall performance and the variability of each method. In addition to internal cross-validation, an independent external cohort was used to further evaluate the generalization ability of the model. The external cohort was excluded from all training and model selection procedures and was used solely for independent testing.

Segmentation performance was comprehensively evaluated using Dice, intersection over union (IoU), 95th percentile Hausdorff distance (HD95), precision, and recall. These metrics respectively quantify region overlap, boundary accuracy, and the trade-off between false positive and false negative predictions. Results are reported as the mean and standard deviation across the five folds unless otherwise noted.

\subsection{Main Results}

\begin{figure*}
    \centering
    \includegraphics[width=\textwidth]{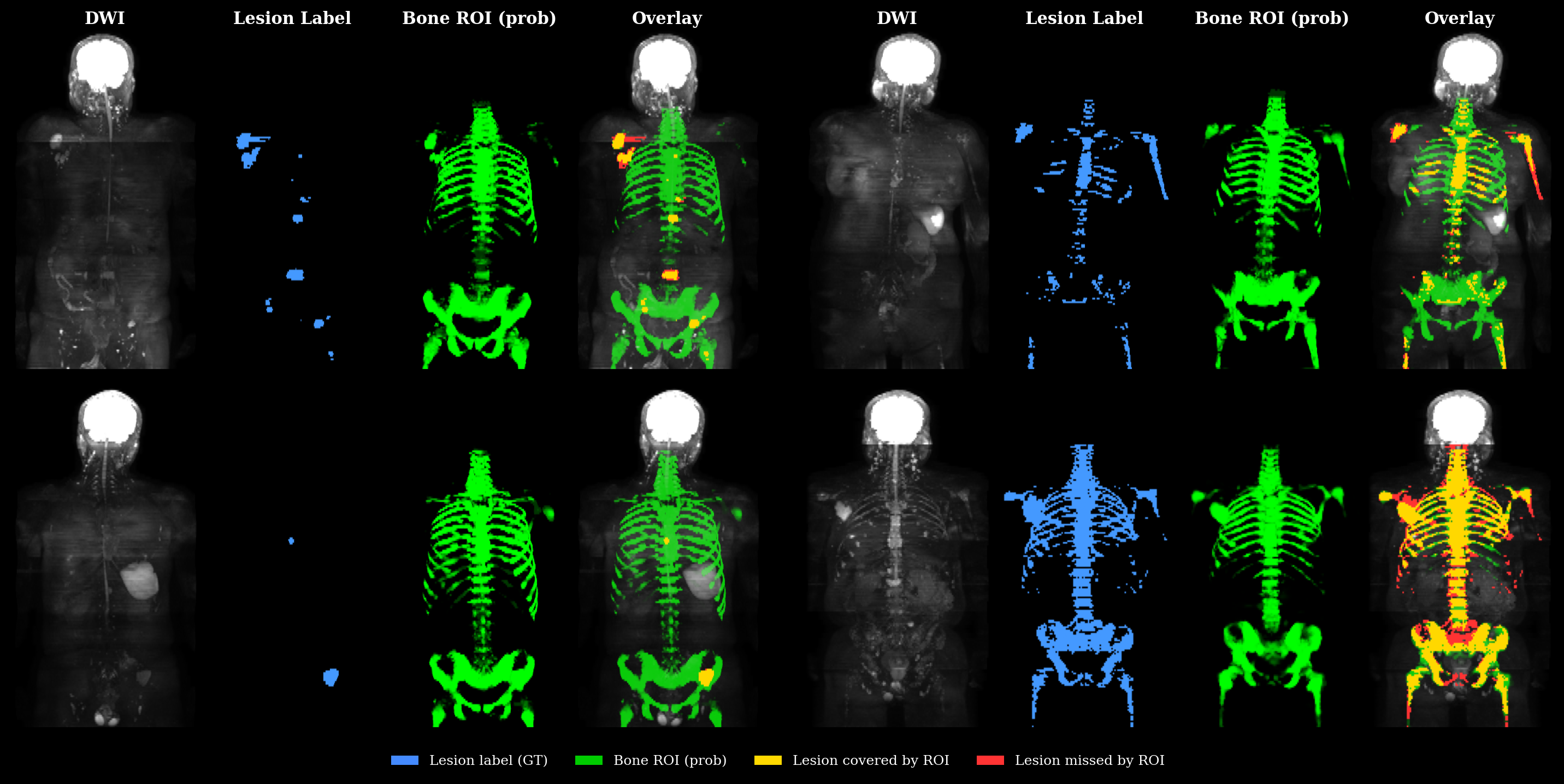}
    \caption{\textbf{Visualization of ADC-based bone ROI generation.} Representative sparse/dense and focal/diffuse cases are shown to illustrate ROI generation across heterogeneous patients. In all cases, the predicted ROI forms anatomically coherent skeletal structures while covering a substantial portion of lesion regions. Although trained with lesion-derived supervision, the ROI generator does not simply fit sparse lesion labels, but instead learns stable bone-related structures for downstream segmentation.}
    \label{fig:roi_generation}
\end{figure*}

\begin{figure*}
    \centering
    \includegraphics[width=\textwidth]{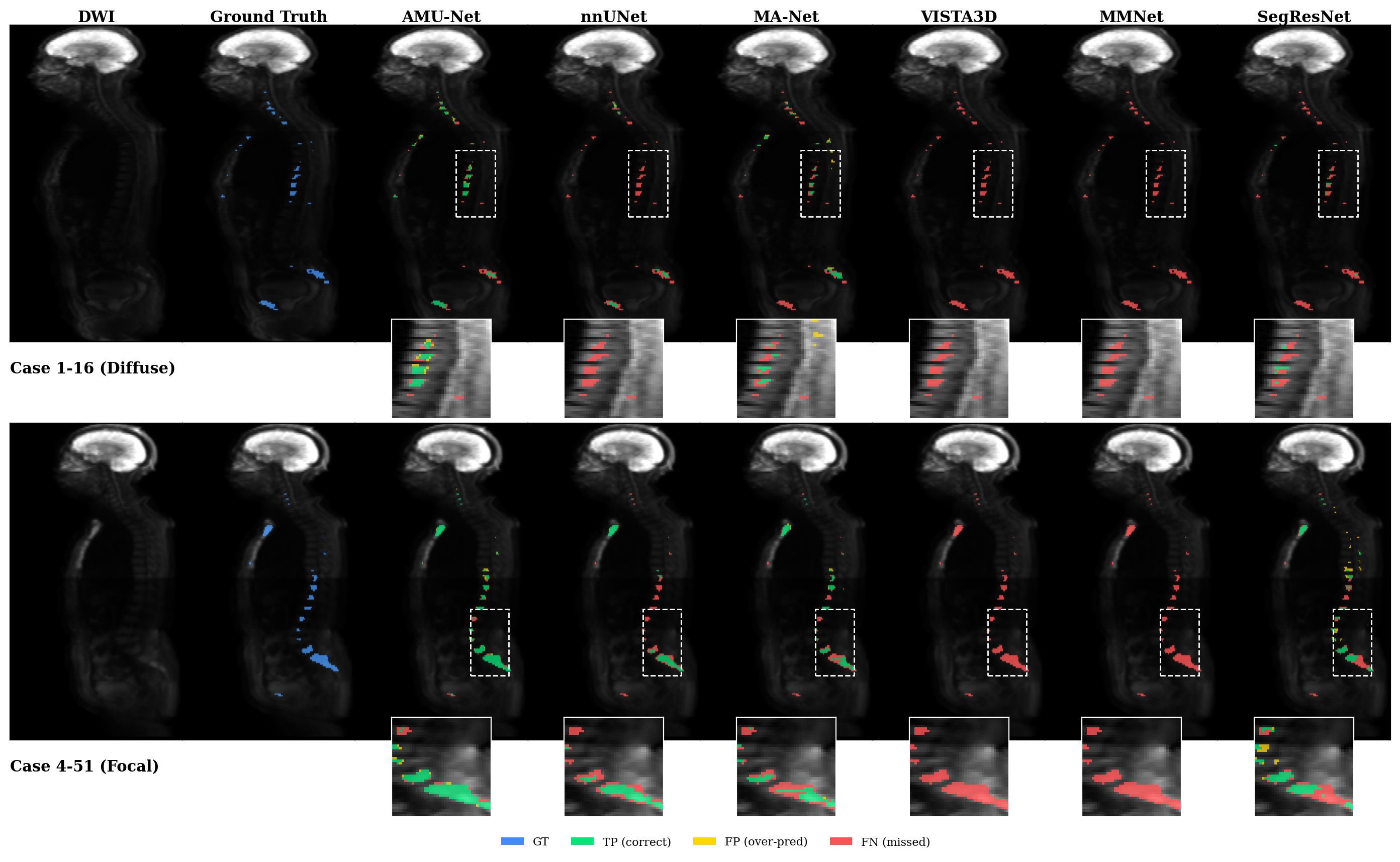}
    \caption{\textbf{Visual comparison of lesion segmentation results.} Representative diffuse and focal cases show that AMU-Net achieves more complete lesion coverage and fewer obvious misses than the compared methods. Zoomed-in views are provided for the boxed regions.}
    \label{fig:visualization_segmentation}
\end{figure*}

\begin{figure*}
    \centering
    \includegraphics[width=\textwidth]{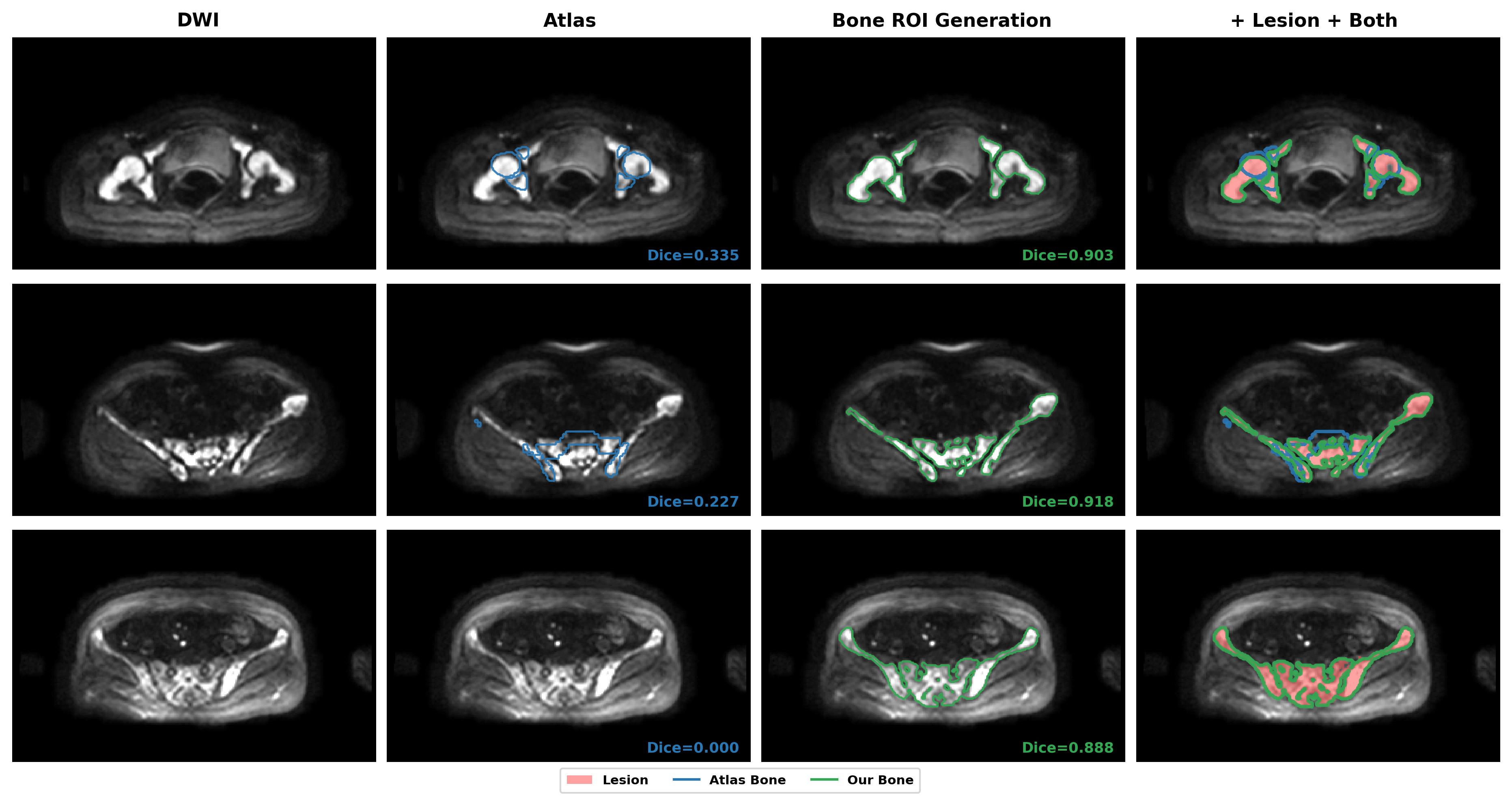}
    \caption{\textbf{Qualitative comparison between a registration pipeline and the proposed direct bone ROI generation method.}The registration-based method often shows spatial mismatch after mapping Dixon-derived bone masks into DWI space, whereas our method produces bone ROIs that better align with lesion-bearing skeletal regions.}
    \label{fig:registraion_vs_generation}
    \vspace{-8pt}
\end{figure*}

\subsubsection{ROI Coverage Analysis}

We first examined whether the generated bone ROI could provide meaningful guidance for lesion segmentation. As shown in Fig.~\ref{fig:roi_generation}, the ADC-based ROI generator produced anatomically coherent pseudo-skeletal structures even in cases with sparse lesion annotations, indicating that it learned a structural bone prior rather than merely fitting lesion labels. Despite the sparsity of the annotations, the generated ROI still covered most lesion regions in both focal and diffuse cases, supporting its role as an effective anatomical guide for downstream segmentation.

\subsubsection{Internal Benchmark Results}

Table~\ref{tab:main_results} summarizes the internal benchmark results. We compared AMU-Net against a diverse set of strong baselines, including representative 3D CNN, 3D transformer, and 2D CNN models, under the same 5-fold evaluation protocol. Among all evaluated methods, AMU-Net achieved the best overall performance, with the highest Dice (76.24$\pm$0.5) and IoU (65.93$\pm$1.0) and the lowest HD95 (10.1$\pm$0.7), indicating superior overlap accuracy and boundary delineation. Notably, nnU-Net was the strongest competing baseline, yet AMU-Net still improved Dice by \textbf{3.1\%} and further reduced HD95. These results demonstrate that the proposed anatomy-guided and ADC-aware design yields consistent gains over a broad range of competitive segmentation architectures.

AMU-Net did not achieve the highest precision or recall individually, but this does not indicate weaker detection ability. In practice, very high precision often reflects a conservative prediction strategy that suppresses false positives at the cost of more missed lesions, whereas very high recall usually corresponds to a more aggressive strategy that captures more lesions but also introduces more false-positive predictions. By achieving the second-best precision and tied second-best recall, AMU-Net showed a better balance between lesion sensitivity and false-positive control. This trend is also consistent with the qualitative comparisons in Fig.~\ref{fig:visualization_segmentation}, where AMU-Net showed more complete lesion coverage with fewer obvious misses.

\begin{table}[t]
\centering
\caption{Ablation results of Stage-II components in AMU-Net. Enc-XAttn: encoder-level ROI-gated cross-attention; Dec-Gate: decoder-level asymmetric gate.}
\label{tab:ablation}
\resizebox{\columnwidth}{!}{%
\begin{tabular}{lccc}
\toprule
Variant & Dice (\%) $\uparrow$ & IoU (\%) $\uparrow$ & HD95 $\downarrow$ \\
\midrule
\multicolumn{4}{l}{\textit{Structural ablation}} \\
Full & 76.24$\pm$0.5 & 65.93$\pm$1.0 & 10.1$\pm$0.7 \\
w/o Enc-XAttn & 73.45$\pm$0.8 & 63.52$\pm$1.1 & 12.5$\pm$0.5 \\
w/o Dec-Gate & 74.87$\pm$0.7 & 64.74$\pm$0.9 & 11.0$\pm$0.5 \\
w/o Both & 73.13$\pm$1.5 & 63.24$\pm$1.3 & 11.8$\pm$1.5 \\
\midrule
\multicolumn{4}{l}{\textit{ROI guidance ablation}} \\
w/o Enc-ROI & 75.80$\pm$0.4 & 65.55$\pm$0.5 & 11.3$\pm$0.5 \\
w/o Dec-ROI & 75.10$\pm$0.6 & 64.94$\pm$0.8 & 11.9$\pm$0.5 \\
\bottomrule
\end{tabular}%
}
\vspace{-8pt}
\end{table}

\subsubsection{External Validation Results}

To further evaluate the generalization capability of AMU-Net, we conducted a zero-shot external validation on the independent cohort collected from Tianjin Institute of Hematology Hospital, without any fine-tuning or domain adaptation. As shown in Table~\ref{tab:external_results}, the overall performance decreased for all methods compared with the internal evaluation, reflecting the substantial distribution shift between institutions. In particular, the external cohort contains a considerably larger proportion of normal cases, whereas only one normal case was available during training. Such a mismatch makes lesion localization substantially more challenging and lowers the overall quantitative metrics across all competing methods.

Despite this challenging setting, AMU-Net consistently achieved the best overall performance, outperforming all baseline methods in Dice and IoU while maintaining competitive Precision and Recall. Notably, nnU-Net, which ranked second on the internal validation, exhibited a much larger performance degradation on the external cohort, whereas AMU-Net retained a clear performance margin. These results suggest that the proposed anatomy-aware gating mechanism, designed according to clinical prior knowledge and lesion distribution characteristics, improves robustness against cross-institutional domain shifts and provides better generalization than purely data-driven segmentation models.

\subsection{Ablation Studies}

\subsubsection{Effect of the Main AMU-Net Components}

Table~\ref{tab:ablation} summarizes the ablation results of the main Stage-II design components in AMU-Net. In the structural ablation, removing either the encoder-level ROI-gated cross-attention or the decoder-level asymmetric gate leads to degraded performance compared with the full model, and removing both modules gives the worst Dice and IoU. This confirms that both components contribute positively to MM lesion segmentation and provide complementary benefits. The larger performance drop caused by removing the encoder module suggests that encoder-level ADC integration plays a particularly important role in supplementing structural information missing from WB-DWI, while the decoder-level asymmetric gate further improves lesion confirmation through DWI--ADC concordance modeling.

\begin{table}[t]
\centering
\caption{Comparison between a registration pipeline and the proposed direct ADC-based bone ROI generation method. Best results are shown in \textbf{bold}.}
\label{tab:bone_roi_compare}
\setlength{\tabcolsep}{6pt}
\renewcommand{\arraystretch}{1.1}
\resizebox{\columnwidth}{!}{
\begin{tabular}{@{}lcc@{}}
\toprule
Metric & Registration Method & Our Generation Method \\
\midrule
Speed per case & $\sim$43.5 s & \textbf{$\sim$4.2 s} \\
Bone--Lesion Dice & $23.04 \pm 13.07$ & \textbf{62.34 $\pm$ 26.96} \\
Lesion recall  & 22.6\% & \textbf{70.9\%} \\
\bottomrule
\end{tabular}
}
\vspace{-8pt}
\end{table}

\subsubsection{Effect of ROI Guidance}

The lower part of Table~\ref{tab:ablation} evaluates the role of ROI guidance within the full model. Removing encoder-side ROI gating or decoder-side ROI amplification both reduces performance, indicating that explicit skeletal prior benefits both structural supplementation and lesion confirmation. This effect is not only attributable to spatial restriction. As shown in Fig.~\ref{fig:adc_two_views}, the average ADC value cannot serve as a reliable indicator of disease across patients, because marrow appearance can vary substantially with individual factors such as age and baseline marrow composition. As a result, lesion assessment is often better understood relative to the patient's own skeletal marrow background rather than by absolute signal level alone. Bone ROI guidance therefore helps the model interpret DWI and ADC patterns within the patient's marrow space, making it easier to distinguish focal lesion-related abnormalities from globally increased or individually variable marrow signal.

\subsection{Practical Value of Direct Bone ROI Generation}

Beyond the final segmentation results, we further examined the practical value of Stage-I bone ROI generation. In routine MM imaging, Dixon is often acquired together with WB-DWI, and bone segmentation on Dixon is known to be highly accurate. We therefore used a Dixon-to-DWI registration pipeline as a strong reference to compare ROI acquisition efficiency and lesion coverage. Despite this favorable reference, the proposed direct ADC-based ROI generation was both substantially faster and more lesion-relevant. It reduced the per-case ROI acquisition time from approximately 43.5 s to 4.2 s, achieving a nearly 10$\times$ speedup, while improving bone--lesion overlap by 39.3 percentage points (Dice: 62.34 vs. 23.04) and lesion recall within the ROI by 48.3 percentage points (70.9\% vs. 22.6\%). It also showed more reliable case-level behavior, with 79.1\% of cases achieving Dice $>$ 0.5, compared with 0\% for the registration-based reference, and no complete failures versus 3.0\%. These results indicate that our ADC-based bone ROI generation method is faster, more accurate, and more practical than the registration-based alternative for whole-body DWI analysis.

\section{Conclusion}

In this work, we presented a two-stage framework for multiple myeloma lesion segmentation on whole-body DWI. As MM is a relatively rare hematologic malignancy, well-annotated WB-DWI data are particularly difficult to acquire, making data-efficient model design especially important. At the same time, although WB-DWI is well suited for visualizing MM lesions, automated segmentation on this modality remains highly challenging because of limited anatomical delineation and the low specificity of marrow hyperintensity. To address these challenges, the proposed framework maximizes the use of available multimodal information by generating bone ROI guidance directly from ADC and performing lesion segmentation with AMU-Net, which leverages clinically motivated DWI--ADC interaction for structural supplementation and lesion confirmation. Despite limited training data, the method achieves clear and consistent performance gains, supporting the value of combining bone-aware guidance with structured multimodal integration for MM lesion segmentation.


\section{Acknowlegments}
This work was supported in part by the National High Level Hospital Clinical Research Funding under Grant 2025-PUMCH-D-002,  in part by the National Natural Science Foundation of China under Grant 82372051, in part by the Peking Union Medical College Hospital Talent Cultivation Program (Category B) under Grant UGG06278, in part by the Beijing Natural Science Foundation under Grant L222099, and in part by Damo Academy through Damo Academy Research Intern Program.



\printcredits

\bibliographystyle{cas-model2-names}

\bibliography{cas-refs}



\end{document}